# DL-PDE: Deep-learning based data-driven discovery of partial differential equations from discrete and noisy data

**Running Title**: Deep-learning based Data-driven PDE Discovery

Hao Xu[a], Haibin Chang[a], and Dongxiao Zhang[b,c*]

[a] BIC-ESAT, ERE, and SKLTCS, College of Engineering, Peking University, Beijing 100871, P. R. China

[b] School of Environmental Science and Engineering, Southern University of Science and Technology, Shenzhen 518055, P. R. China

[c] Intelligent Energy Lab, Peng Cheng Laboratory, Shenzhen 518000, P. R. China

[*] Corresponding author

E-mail address: 390260267@pku.edu.cn (H. Xu); changhaibin@pku.edu.cn (H. Chang); zhangdx@sustech.edu.cn (D. Zhang)

**Abstract**: In recent years, data-driven methods have been developed to learn dynamical systems and partial differential equations (PDE). The goal of such work is discovering unknown physics and the corresponding equations. However, prior to achieving this goal, major challenges remain to be resolved, including learning PDE under noisy data and limited discrete data. To overcome these challenges, in this work, a deep-learning based data-driven method, called DL-PDE, is developed to discover the governing PDEs of underlying physical processes. The DL-PDE method combines deep learning via neural networks and data-driven discovery of PDE via sparse regressions. In the DL-PDE, a neural network is first trained, and then a large amount of meta-data is generated, and the required derivatives are calculated by automatic differentiation. Finally, the form of PDE is discovered by sparse regression. The proposed method is tested with physical processes, governed by groundwater flow equation, convection-diffusion equation, Burgers equation and Korteweg–de Vries (KdV) equation, for proof-of-concept and applications in real-world engineering settings. The proposed method achieves satisfactory results when data are noisy and limited.

# I. INTRODUCTION

As data acquisition and storage ability have increased, data-driven methods have been utilized for solving various problems in different fields. In recent years, data-driven discovery of governing equations of physical problems has attracted much attention. Instead of building models from physical laws, the goal of such an approach is to discover unknown physics and the corresponding equations directly from limited observation data. Substantial progress has been made in terms of proof-of-concept and preliminary applications. Among these investigations, sparse regression methods are frequently used techniques, which show promise for discovering the governing partial differential equations (PDEs) of various problems. Using sparse regression aims to identify a small number of terms that constitute a governing equation from a predefined large candidate library, and a parsimonious model can usually be obtained. Sparse identification of nonlinear dynamics (SINDy), sequential threshold ridge regression (STRidge), and Lasso are proposed to identify PDE from data [1–3]. Since then, a large body of extant literature has investigated data-driven discovery of governing equations using sparse regression [4–16]. Despite the numerous successes achieved with sparse regression-based methods, major challenges remain when faced with noisy data and limited data. Since numerical approximation of derivatives is requisite in these methods, the results may be unstable and ill-conditioned when handling noisy data [17]. Total variation, polynomial interpolation, and the integral form are utilized to handle noisy data [1,2,9]. However, these strategies can only lessen the difficulties associated with noisy data to a certain extent.

Besides the sparse regression method, other techniques, such as Gaussian process and neural networks, are also used for performing data-driven discovery of governing equations. For example, Raissi et al. [18] proposed a framework that utilizes the Gaussian process to discover governing equations. In their proposed framework, the parameters of the differential operator are turned into hyper-parameters of some covariance functions, and are learned by the maximum likelihood method. Meanwhile, physics-informed neural networks (PINN) is presented for solving forward and inverse problems of PDE [19]. In the PINN, by adding a PDE constraint term in the loss function, in addition to the data match term, the accuracy of the results can be improved, and the coefficients of the PDE terms can be learned. Avoiding the numerical approximation of derivatives, both the Gaussian process-based method and the neural network-based method require less data and are less sensitive to data noise. However, in the above-mentioned works, the PDE of the considered problem is supposed to have a known structure, and only the coefficients of the PDE terms are learned from data, which limits its application for PDE discovery. To overcome this limitation, Raissi [20] modified the PINN by introducing two neural networks for approximating the unknown solution, as well as the unknown PDE. Even though this modification enables the PINN to solve problems with unknown PDE structures, the learned neural network approximation of the unknown PDE is a black box, and thus lacks interpretability. Long et al. [21] employed a convolutional neural network to identify the form of the unknown PDE. However, parsimony of the results may not be guaranteed. An algorithm for discovering PDE from a complex dataset was proposed by Berg and Nystrom [22]. In their work, derivatives are calculated by automatic differentiation of neural network. Their method, however, requires a large amount of observation data (e.g., over 250000 data) and is only robust to a relatively low noise level.

For discovery of governing equations, a qualified data-driven method should obtain an interpretable, parsimonious model with high accuracy, and it should be insensitive to data noise. Learning from the advantages and disadvantages of the existing methods, in this work, we propose a new data-driven method, called DL-PDE, which combines deep neural network and sparse regression method for discovery of PDE. In DL-PDE, a neural network is first trained to well learn (fit) the data. Then, the trained neural network is utilized to generate meta-data and calculate corresponding derivatives. Finally, sparse regression is used for PDE discovery. The proposed DL-PDE method inherits the properties of both the neural network method and the sparse regression method, and thus can obtain a parsimonious model and be insensitive to data noise. Four PDEs, including groundwater flow equation, convection-diffusion equation, Burgers equation and Korteweg–de Vries (KdV) equation, are employed for testing the proposed method. The influence of noisy data and limited discrete data are investigated, and satisfactory results are obtained.

## II. METHODOLOGY

### A. PDE discovery

In this work, we aim to investigate the data-driven discovery of PDEs with the following form:

$$u_t = \Phi(u) \cdot \xi, \tag{1}$$

with

$$\Phi(u) = [1, u, u^2, u_x, u_{xx}, ..., uu_x, uu_{xx}, ...]. \tag{2}$$

where $u$ denotes the solution of the considered problem; $\Phi(u)$ denotes the candidate library of potential PDE terms; and $\xi$ denotes the coefficient vector. In this work, $\Phi(u)$ is supposed to be sufficiently rich, which means that the terms that constitute the PDE of the considered problem are contained in $\Phi(u)$.

Considering that a PDE usually consists of a small number of terms, data-driven discovery of a PDE aims to find a sparse coefficient. In order to learn the coefficient, spatial and temporal observation data are requisite. Here, observation data (or meta-data) are denoted as $\{u(x_i, t_i)\}_{i=1}^{N}$. Since the PDE holds for each data point, we have:

$$\begin{bmatrix} u_t(x_1,t_1) \\ u_t(x_2,t_2) \\ \vdots \\ u_t(x_N,t_N) \end{bmatrix} = \begin{bmatrix} 1 & u(x_1,t_1) & \cdots & uu_{xx}(x_1,t_1) & \cdots \\ 1 & u(x_2,t_2) & \cdots & uu_{xx}(x_2,t_2) & \cdots \\ \vdots & \vdots & \ddots & \vdots & \ddots \\ 1 & u(x_N,t_N) & \cdots & uu_{xx}(x_N,t_N) & \cdots \end{bmatrix} \cdot \xi, \tag{3}$$

which can be rewritten as:

$$U_t = \Psi(U) \cdot \xi \tag{4}$$

where $\Psi(U)$ denotes the discretized form of the candidate library. It is worth noting that, for learning $\xi$, $U_t$ and $\Psi(U)$ should be prepared beforehand, and $\Psi(U)$ usually contains different orders of derivatives of $u$ with respect to the spatial variable at all data points. Therefore, calculating the derivatives contained in $U_t$ and $\Psi(U)$ is necessary for learning $\xi$. Numerical approximation of derivatives from data is straightforward, for example, by using the finite difference method. However, the results may be unstable and ill-conditioned when dealing with noisy data [17]. In this work, the required derivatives are obtained by using automatic differentiation of the neural network that is trained with available data.

## B. Neural network

In this work, a neural network is utilized to approximate the physical problem solution and obtain the required derivatives in Eq. (4). Here, we first briefly introduce the neural network approximation. A feed forward fully-connected neural network is utilized in this work, and its structure is shown in Fig. 1. The neural network comprises an input layer, an output layer, and one or several layer(s) between the input and output layers that are termed hidden layer(s). Each hidden layer is composed of multiple neurons. Two adjacent layers are connected as follows:

$$\mathbf{z}_l = \sigma(\mathbf{W}_l \mathbf{z}_{l-1} + \mathbf{b}_l), \, l = 1, \ldots, L-1 \tag{5}$$

where $l$ denotes the layer index; $\mathbf{W}$ denotes the weight matrix; $\mathbf{b}$ denotes the bias vector; and $\sigma$ denotes the activation function. Consequently, using a neural network approximation, the relationship between the input vector $\mathbf{z}_0$ and output prediction $\mathbf{z}_L$ can be expressed as:

$$\mathbf{z}_L = NN(\mathbf{z}_0; \theta) = \mathbf{W}_L \sigma(\cdots \sigma(W_2 \sigma(W_1 \mathbf{z}_0 + \mathbf{b}_1) + \mathbf{b}_2)) + \mathbf{b}_L, \tag{6}$$

where $\theta$ denotes the collection of all learnable coefficients, which can be written as:

$$\theta = \{W_1, b_1, W_2, b_2, ..., W_L, b_L\}. \tag{7}$$

For approximating the solution of a physical problem, the input vector comprises the spatial and temporal variable, which is $\mathbf{z}_0 = [x, t]^T$; and the output prediction is a scalar $u(x,t)$. Suppose that there are $N$ observation data, $\{u(x_i, t_i)\}_{i=1}^N$. In order to train the neural network, a loss function is then defined as follows:

$$Loss(\theta) = \sum_{i=1}^{N} [u(x_i, t_i) - NN(x_i, t_i; \theta)]^2. \tag{8}$$

In this work, the Adam optimizer is utilized to minimize the loss function for training the neural network [23].

After training the neural network with available observations from the underlying physical problem, the required derivatives can be easily accessed by applying automatic differentiation. Different from numerical differentiation, automatic differentiation possesses the advantages of small influence of noise, desirable stability, and good expandability. In addition to observation data, a large amount of meta-data can also be generated using the trained neural network, which is vital to the accuracy of PDE discovery with limited and noisy data.

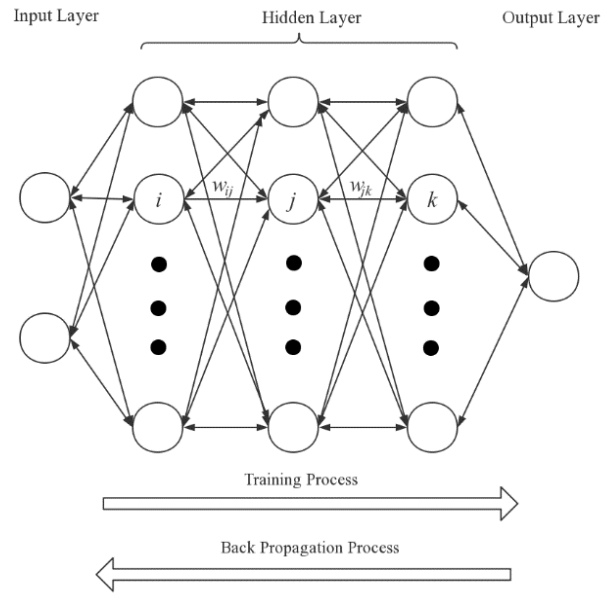

**FIG. 1.** The structure of a feed forward fully-connected neural network. The forward arrow indicates the training process of the neural network, and the backward arrow indicates the back propagation process of the neural network.

## C. DL-PDE

In this work, we propose a new data-driven method, called DL-PDE, which combines deep neural network and sparse regression method for discovering PDE. For a physical problem, a deep neural network is trained using available data. According to the universal approximation theorem of neural network, functions of any complexity can be approximated by a neural network with any precision [24]. Next, similar to that in sparse regression-based methods, a candidate library of potential PDE terms is designed. Automatic differentiation of neural networks is then utilized to calculate the derivatives, which offers the advantages of small influence of noise, good stability, and desirable expandability. Different from the previous work by Berg and Nystrom [22], in DL-PDE a large amount of meta-data and their derivatives can be generated using the trained neural network, instead of the derivatives of the actual observation data being calculated by the neural network. When the amount of observation data is small, the generation of meta-data is critical in accurately discovering the correct equations. Finally, sparse regression methods are adopted to identify the sparse terms from the candidate library that constitute a PDE. It is worth noting that, although STRidge [2] is found suitable and thus adopted in this study, other sparse regression methods, such as Lasso [3] and sparse Bayesian inference [14], can also be employed for this purpose. The workflow of DL-PDE is shown in Fig. 2.

Here, we briefly introduce STRidge, additional details of which can be found in Rudy et al. [2]. Solving Eq. (4) using ridge regression can be achieved by using the following formula:

$$\begin{aligned}\hat{\xi} &= \arg\min_{\xi} \left\|\Theta\xi - U_t\right\|_2^2 + \lambda\left\|\xi\right\|_2^2 \\ &= (\Theta^T\Theta + \lambda I)^{-1}\Theta^T U_t.\end{aligned} \tag{9}$$

In order to obtain sparse results, an appropriate threshold *tol* is introduced to select coefficients. The coefficients that are larger than *tol* are retained, while the coefficients that are smaller than *tol* are omitted. This process will continue with the remaining terms until the number of terms no longer changes.

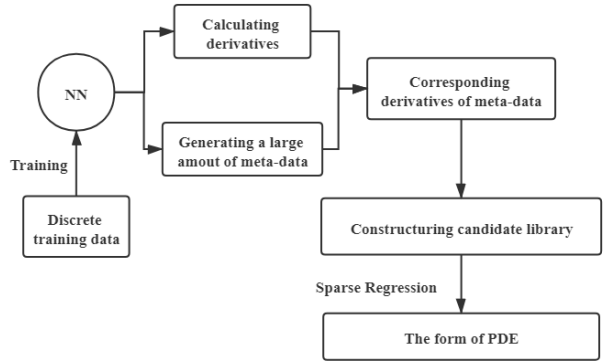

**FIG. 2**. The workflow of DL-PDE.

**III. Results**

In this section, we use some classical physical processes described by groundwater flow equation, convection-diffusion equation, Burgers equation, and KdV equation to test the performance of the DL-PDE method. In this work, a nine-layer deep neural network with 20 neurons per hidden layer is utilized and activation functions are tanh(x), except for the case of KdV equation.

### A. PDE discovery with limited and noisy data

#### 1. Learning groundwater flow equation

Firstly, DL-PDE is utilized to learn the governing equation of groundwater flow in a saturated confined aquifer. For proof-of-concept, the data are generated through numerical simulation. We consider one dimensional (1-D) saturated flow, whose governing equation is as follows:

$$Kh_{xx} = S_s h_t \qquad (10)$$

In this case, the conductivity $K$ is supposed to be homogeneous with a value of 1 m/d, and the specific storage $S_s$ is 0.01 m$^{-1}$. For the dataset, $x \in [0,1010]$, $t \in [0,200]$, the number of spatial observation points $n_x$ =101 with $\Delta x = 10$, the number of temporal observation points $n_t$ =100, and the size of the data $N_d$ =10100. Since $S_s$ is always very small in the groundwater flow equation, we will pre-process the data prior to proceeding with the algorithm. We use $x^* = \dfrac{x}{\Delta x}$, $t^* = t$, and $h^* = \Delta x \cdot h$ to replace $x, t,$ and $h$, respectively, and the equation is converted into:

$$h_{t^*}^* = \lambda h_{x^* x^*}^* \qquad (11)$$

In this case, $\lambda$ =1. The candidate library is constructed as:

$$\Phi = [1 \quad h_{x^*}^* \quad h_{x^* x^*}^* \quad h_{x^* x^* x^*}^* \quad h^* h_{x^*}^* \quad h^* h_{x^* x^*}^* \quad h^* h_{x^* x^* x^*}^* \quad h^{*2} h_{x^*}^* \quad h^{*2} h_{x^* x^*}^* \quad h^{*2} h_{x^* x^* x^*}^*] \qquad (12)$$

The candidate library is a sufficient library with four linear terms and six nonlinear terms. Here, we consider the derivatives up to order three. Meta-data are generated in the domain $x^{*'} \in [4,105]$ and $t^{*'} \in [0,200]$ with the number of spatial observation points $n'_x$ =101, the number of temporal observation points $n'_t$ =1000, and the size of the data set $N'_d$ =101000. Its data volume is 10 times the original data volume, which means that much more meta-data are generated. This is crucial for improving the performance of the sparse regression. Finally, we utilize the derivatives of meta-data to establish a candidate library from Eq. (12), and perform STRidge to obtain the form of the equation and corresponding coefficients.

The performance of DL-PDE is investigated under different data volumes. 2500 (24.8%), 1000 (9.9%), 500 (4.95%), 200 (2.48%), and 100 (1.24%) data are randomly chosen to form new datasets, respectively. Illustrations of selected data are shown in Appendix A. Then, DL-PDE is performed to learn PDE correspondingly. The results are summarized in Table I.

**TABLE I.** Summary of PDEs learned using DL-PDE for the case of groundwater flow with different data volume.

| Data Volume | Learned Equation |
|---|---|
| **Correct PDE** | $h^*_{t^*} = h^*_{x^*x^*}$ |
| **2500 data** | $h^*_{t^*} = 0.992 h^*_{x^*x^*}$ |
| **500 data** | $h^*_{t^*} = 0.988 h^*_{x^*x^*}$ |
| **200 data** | $h^*_{t^*} = 0.987 h^*_{x^*x^*}$ |
| **100 data** | $h^*_{t^*} = 0.731 h^*_{x^*x^*}$ |

From Table I, it can be seen that the performance of DL-PDE increases as the amount of training data increases. To further quantify the accuracy of the learned PDE compared to the true underlying equation, a relative error is defined as:

$$\text{relative error} = \frac{|u(x,t) - u'(x,t)|}{\max(u(x,t)) - \min(u(x,t))} \times 100\% \qquad (13)$$

where $u(x,t)$ is the solution of the true PDE; $u'(x,t)$ is the solution of the learned PDE; and $\max(u(x,t))$ and $\min(u(x,t))$ are the maximum and minimum value of $u(x,t)$, respectively. The relative error is shown in Fig. 3. It can be seen that, even in the case of small data volume, such as 500 data and 200 data, DL-PDE can still learn the PDE form accurately with no more than 0.5% error. When the data volume is too small, such as 100 data, although the correct PDE form can be learned, the error is large.

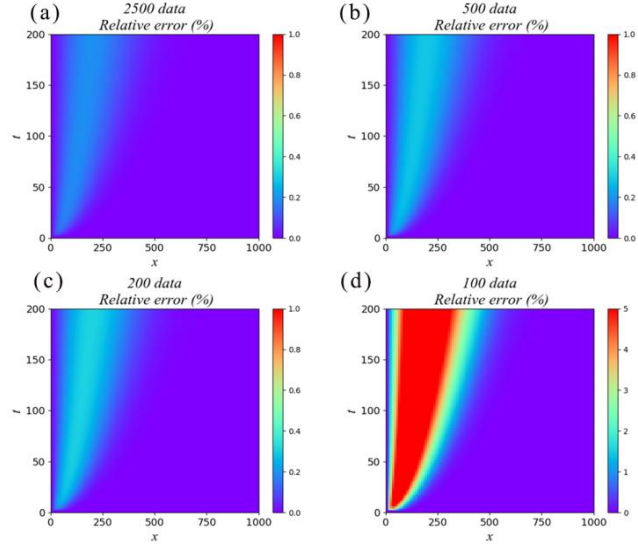

**FIG. 3.** Relative error of the learned PDEs with different data volume for the groundwater flow equation. (a) 2500 data, (b) 500 data, (c) 200 data, and (d) 100 data. The maximum error in the color bar of (a), (b) and (c) is 1%, while the maximum error in the color bar of (d) is 5%.

Next, the performance of DL-PDE is tested with noisy data. In this work, noise is synthetically added to the data at each monitoring location as follows:

$$u(x,t) = u(x,t) \times (1 + \delta \times e) \quad (14)$$

where $\delta$ denotes the noise level; and $e$ denotes the uniform random variable, taking values from -1 to 1 [8]. We randomly select 2500 data to train the neural network. Two noise levels, 1% and 5%, are added to the data. The results are summarized in Table II. The relative error is shown in Fig. 4.

**TABLE II.** Summary of PDEs learned using DL-PDE for the case of groundwater flow with noisy data.

| Noise Level | Learned Equation |
|---|---|
| **Correct PDE** | $h^*_{t^*} = h^*_{x^*x^*}$ |
| **Clean data** | $h^*_{t^*} = 0.992 h^*_{x^*x^*}$ |
| **1% noise** | $h^*_{t^*} = 0.971 h^*_{x^*x^*}$ |
| **5% noise** | $h^*_{t^*} = 0.906 h^*_{x^*x^*}$ |

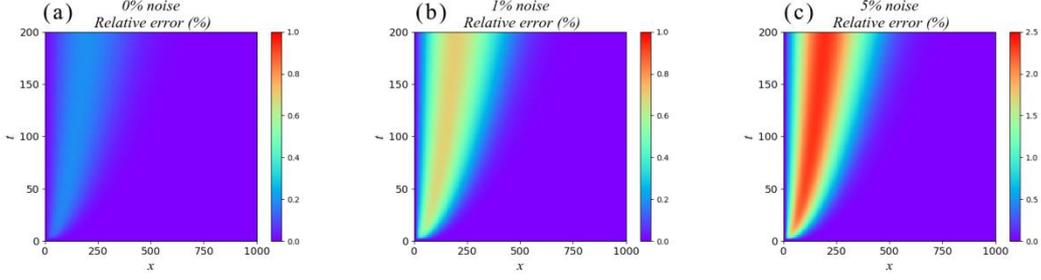

**FIG. 4.** Relative error of the learned PDEs with different noise level for the groundwater flow equation. (a) clean data, (b) 1% noise, and (c) 5% noise. The maximum error in the color bar of (a) and (b) is 1%, while the maximum error in the color bar of (c) is 2.5%.

From Table II, it can be seen that DL-PDE is robust to noise. For the groundwater flow equation, an earlier work [8] found robust results with 5% noise, as well, but with noisy data smoothed and all of the data used. In the present study, although noisy data are not smoothed, DL-PDE can still obtain the PDE form and the corresponding coefficients with only 25% of total data.

*2. Learning convection-diffusion equation*

The convection-diffusion equation is considered next, whose governing equation reads as:

$$C_t = -v_x C_x + D_L C_{xx} \qquad (15)$$

To obtain a set of training data, we simulate the convection-diffusion Eq. (15) by numerical simulation. $v_x$ and $D_L$ are set to be 1 and 0.25, respectively. For the dataset, $x \in [0,30]$, $t \in [0,15]$, $n_x = 120$, $n_t = 150$, and $N_d = 18000$. Meta-data are generated in the domain $x' \in [3,15]$ and $t' \in [0,9]$ with $n'_x = 120$, $n'_t = 900$, and $N'_d = 108000$. The candidate library is established as in Eq. (12).

The performance of DL-PDE is tested under different data volume. From the origin dataset, 4000 (22.2%), 2000 (11.10%), 1000 (5.56%), 500 (2.78%), 200 (1.39%), and 100 (0.70%) data are randomly selected, respectively. Then, DL-PDE is utilized to learn PDE correspondingly. The results are summarized in Table III. The relative error is shown in Fig. 5. It can be seen that DL-PDE still works well in the case of small data volume.

Next, different levels of noise are added to the data. 500 data are randomly selected, and DL-PDE is performed. The results are presented in Table IV. The relative error is shown in Fig. 6. For the convection-diffusion equation, a previous investigation reported that it is robust to 10% noise, but noisy data are smoothed and all of the data are used [8]. In contrast, DL-PDE performs very well under noisy data, and is robust to 10% noise even with only 500 data, which accounts for 2.78% of total data. Meanwhile, the relative error is no more than 1%.

**TABLE III.** Summary of PDEs learned using DL-PDE for the case of convection-diffusion with different data volume.

| Data Volume | Learned Equation |
|---|---|
| Correct PDE | $C_t = -C_x + 0.25 C_{xx}$ |
| 4000 data | $C_t = -0.999 C_x + 0.251 C_{xx}$ |
| 1000 data | $C_t = -0.999 C_x + 0.248 C_{xx}$ |
| 500 data | $C_t = -1.000 C_x + 0.248 C_{xx}$ |
| 200 data | $C_t = -0.989 C_x + 0.246 C_{xx}$ |
| 100 data | $C_t = -1.005 C_x + 0.239 C_{xx}$ |

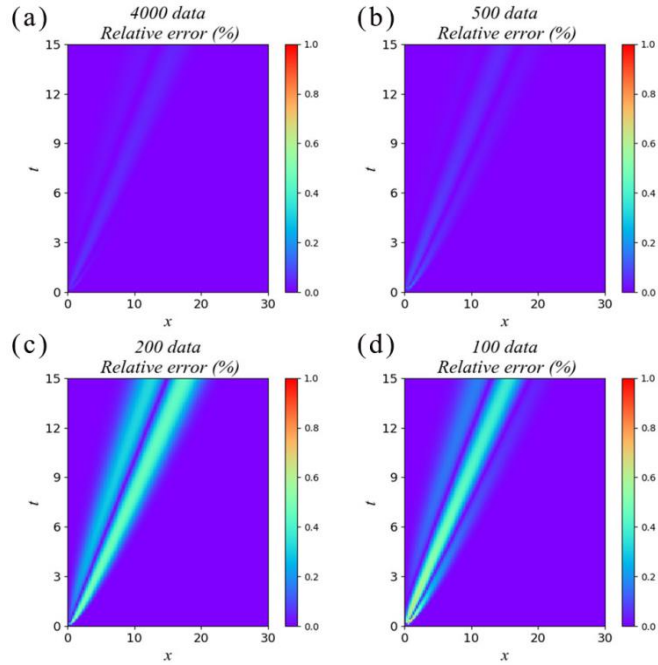

**FIG. 5.** Relative error of learned PDEs with different data volume for convection-diffusion equation. (a) 4000 data, (b) 500 data, (c) 200 data, and (d) 100 data. The maximum error in the color bar is 1%.

**TABLE IV.** Summary of PDEs learned using DL-PDE for the case of convection-diffusion equation with noisy data.

| Noise Level | Learned Equation |
|---|---|
| **Correct PDE** | $C_t = -C_x + 0.25 C_{xx}$ |
| **Clean data** | $C_t = -1.000 C_x + 0.248 C_{xx}$ |
| **1% noise** | $C_t = -0.996 C_x + 0.248 C_{xx}$ |
| **5% noise** | $C_t = -0.995 C_x + 0.247 C_{xx}$ |
| **10% noise** | $C_t = -0.999 C_x + 0.239 C_{xx}$ |

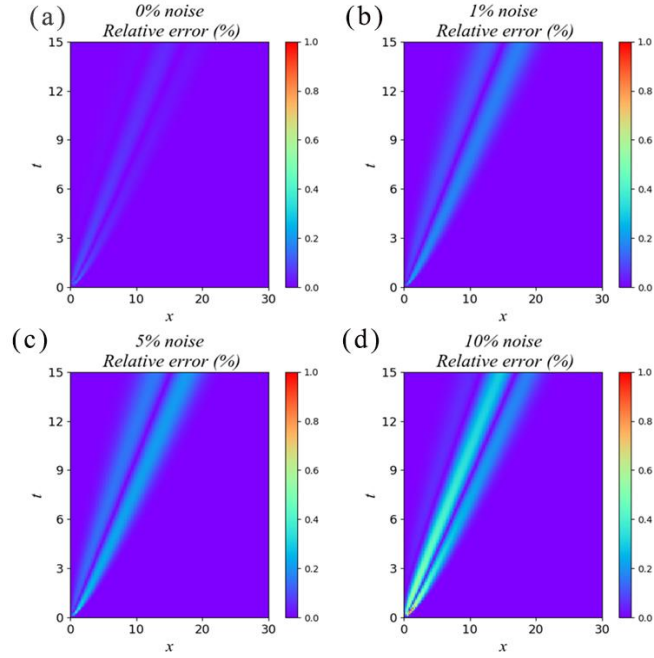

**FIG. 6.** Relative error of learned PDEs with different noise level for convection-diffusion equation. (a) clean data, (b) 1% noise, (c) 5% noise, and (d) 10% noise. The maximum error in the color bar is 1%.

### 3. Learning Burgers equation

One-dimensional Burgers equation can be written as follows:

$$u_t = -uu_x + au_{xx} \tag{16}$$

where $a$ is the diffusion coefficient, and we set $a$=0.1 in this example. Compared with the previous cases, the Burgers equation has a nonlinear term, and thus it is more difficult to learn the equation form. It is used to test the performance of DL-PDE when learning nonlinear terms. For the dataset, $x \in [-8,8)$ and $t \in [0,10]$, $n_x$=256, $n_t$=201, and $N_d$=51456. Meta-data are generated in the domain $x' \in [-8,8)$ and $t' \in [0,9]$ with $n'_x$=320, $n'_t$=180, and $N'_d$=57600. The candidate library is established as in Eq. (12).

DL-PDE is utilized to learn PDEs with different data volume, and the results are summarized in Table V. Next, noise is added to the set of 3000 randomly selected data, and the DL-PDE results are presented in Table VI. For both, the relative error is shown in Fig. 7. For the Burgers equation, an earlier work found that it is robust to 1% noise when noisy data are smoothed and all of the data are used [2]. In contrast, DL-PDE is found to be robust to 5% noise, even in the case of small data volumes with only 1% relative error.

**TABLE V.** Summary of PDEs learned using DL-PDE for the case of the Burgers equation with different data volume.

| Data Volume | Learned Equation |
|---|---|
| **Correct PDE** | $u_t = -uu_x + 0.1u_{xx}$ |
| **3000 data** | $u_t = -0.996uu_x + 0.099u_{xx}$ |
| **2000 data** | $u_t = -0.989uu_x + 0.097u_{xx}$ |
| **1000 data** | $u_t = -0.970uu_x + 0.091u_{xx}$ |

**TABLE VI.** Summary of PDEs learned using DL-PDE for the case of the Burgers equation with noisy data.

| Noise Level | Learned Equation |
|---|---|
| Correct PDE | $u_t = -uu_x + 0.1u_{xx}$ |
| Clean data | $u_t = -0.996uu_x + 0.099u_{xx}$ |
| 1% noise | $u_t = -0.993uu_x + 0.098u_{xx}$ |
| 5% noise | $u_t = -0.986uu_x + 0.095u_{xx}$ |

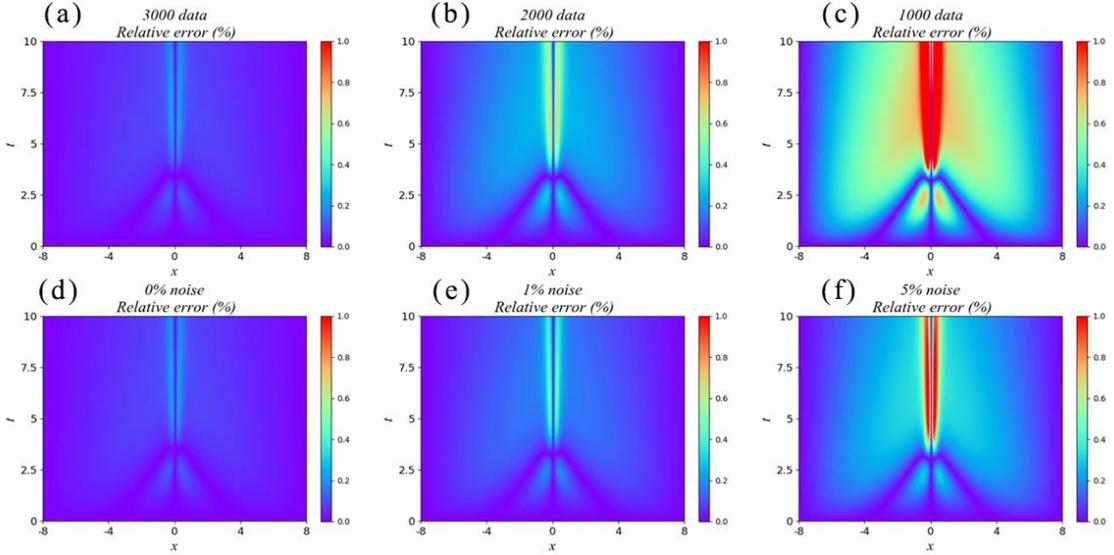

**FIG. 7.** Relative error of learned PDEs for Burgers equation with different data volume: (a) 3000 data, (b) 2000 data, and (c) 1000 data; at different noise level (with 3000 data): (d) clean data, (e) 1% noise, and (f) 5% noise. The maximum error in the color bar is 1%.

*4. Learning the KdV equation*

Finally, we investigate a more complicated equation, the KdV equation, which reads as follows:

$$u_t = -uu_x - bu_{xxx} \tag{17}$$

where *b* is a constant, which is set as *b*=0.0025 in this example. Apparently, a higher-order differential term appears in the equation, which poses a challenge to the accuracy of calculating derivatives. In previous works, the finite difference method is used to calculate derivatives. However, for higher-order differential terms, the error of the numerical difference method may become non-negligible and even affect the PDE-learning process, leading to learning incorrect equation forms. Therefore, the performance of DL-PDE is tested in the presence of high-order differential terms. For the dataset, $x \in [-1,1]$, $t \in [0,1]$, $n_x$=512, $n_t$=201, and $N_d$=102912.

Different from the previous examples, a five-layer deep neural network with 50 neurons per hidden layer is utilized to represent the solution $u$. Activation functions are changed to sin(x) in this case. Meta-data and corresponding derivatives are generated by the trained neural network. Meta-data are generated in the domain $x' \in [-0.5, 0.5]$, and $t' \in [0,1]$ with $n'_x$=1000, $n'_t$=200, and $N'_d$=200000. The candidate library is established as in Eq. (12).

25000 data (24.3% of total data) are randomly selected to train the neural network, and DL-PDE is utilized to learn the PDE. The learned PDEs are given in Table VII, and the corresponding relative error is shown in Fig. 8. It can be seen that DL-PDE can still find the correct equation and corresponding coefficients with high accuracy in the presence of high-order differential terms. For the KdV equation, an earlier work reported that it is robust to 1% noise when noisy data are smoothed and all of the data are used [2]. In contrast, DL-PDE is robust to 10% noise, even with a fraction of data. Meanwhile, from Fig. 8, it can be seen that compared with previous examples, error of learned PDEs is higher, which is nearly 3% with no noise and 5% with 1% noise. In addition, as there are high-order differential terms in the KdV equation, more data are needed to enhance the accuracy of the PDE-learning process. Therefore, for the KdV equation, we do not discuss the performance of the DL-PDE with a smaller data volume.

**TABLE VII.** Summary of PDE learned using DL-PDE for the case of the KdV equation.

| Noise Level | Learned Equation |
|---|---|
| Correct PDE | $u_t = -uu_x - 0.0025u_{xxx}$ |
| Clean data | $u_t = -0.993uu_x - 0.00248u_{xxx}$ |
| 1% noise | $u_t = -0.986uu_x - 0.00247u_{xxx}$ |
| 5% noise | $u_t = -0.968uu_x - 0.00242u_{xxx}$ |
| 10% noise | $u_t = -0.940uu_x - 0.00234u_{xxx}$ |

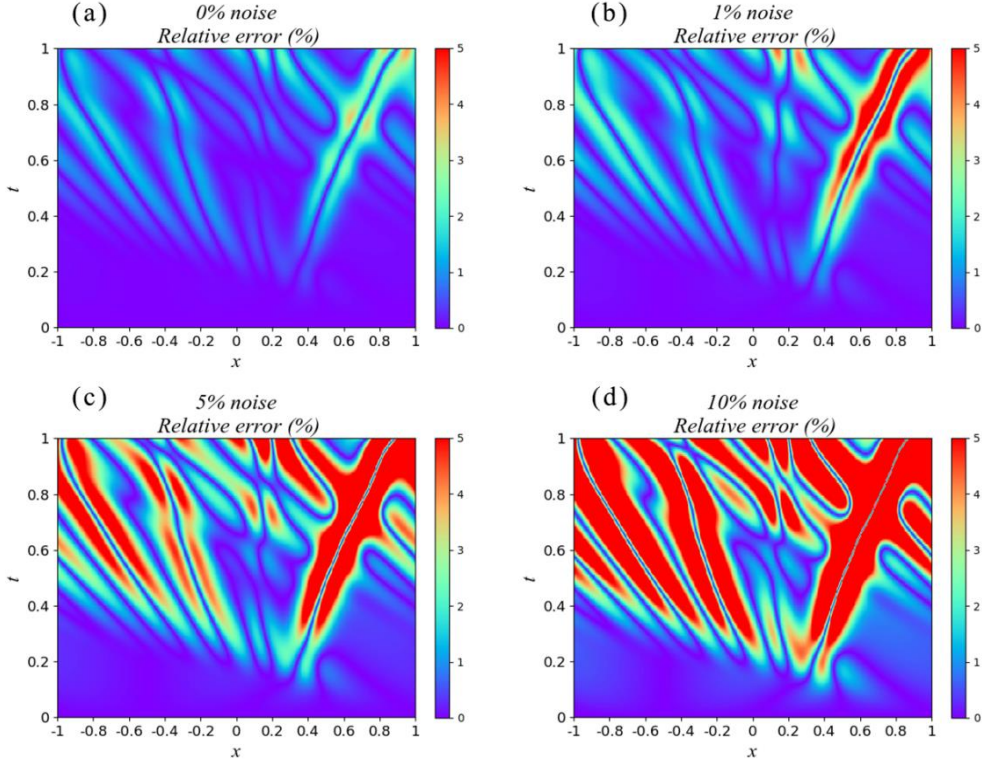

**FIG. 8.** Relative error of learned PDEs with different noise level for the KdV equation. (a) clean data, (b) 1% noise, (c) 5% noise, and (d) 10% noise. The maximum error in the color bar is 5%.

B. Learning PDE in engineering settings

To further examine the performance of the proposed algorithm, DL-PDE is utilized to learn PDE in engineering settings. In previous experiments, observation data points are randomly chosen. In actual engineering settings, however, we may not randomly record data points. In general, there are two ways to record engineering data. The first is to take temporal observations at some limited fixed spatial monitoring locations. For example, in an oil field, a limited number of wells are drilled at fixed locations, from which physical quantities (e.g., production or injection rates) are recorded in time. The second way is to sweep through a space at limited fixed time intervals. For example, an environmental monitoring vehicle passes through an area to measure the concentration of pollutants in a short period of time, or a satellite may cover a certain area to make spatial observations over a fixed time schedule. Therefore, we may either have a time series of observations at fixed locations $x$ or spatial (continuous or discrete) observations at fixed temporal data points $t$. In this subsection, we use DL-PDE to learn PDE in order to test the performance of this algorithm in engineering settings.

Convection-diffusion equation and Burgers equation are considered in this part. Here, fixed temporal observation points is taken as an example. For the convection-diffusion equation, a total of 15 temporal observation points is uniformly taken in the entire domain. For the Burgers equation, a total of 40 temporal observation points is uniformly taken from $t$=0.05 to $t$=9.95. We select the data at all spatial points at these temporal points as the dataset. Selected data are shown in Fig. 9.

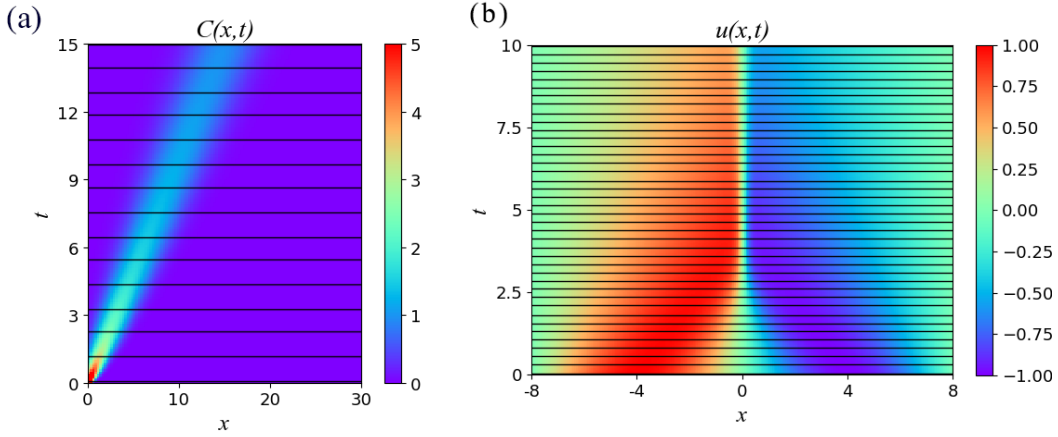

**FIG. 9.** Generating data from fixed time observation points. Training data are generated from 15 temporal observation points for the case of convection-diffusion equation (a) and from 40 temporal observation points for the case of Burgers equation (b). The background represents the solution $u$ in the dataset by heat map, and the black lines are selected data.

A five-layer deep neural network with 50 neurons per hidden layer is utilized to represent the solution $u$, and the activation functions used are sin(x). Meta-data are generated in the same way as in the previous corresponding examples. The results are presented in Table VIII.

TABLE VIII. Summary of PDEs learned using DL-PDE for the case of fixed temporal observation data.

| Noise Level | Learned Equation (convection-diffusion equation) | Learned Equation (Burgers equation) |
| --- | --- | --- |
| Correct PDE | $C_t = -C_x + 0.25 C_{xx}$ | $u_t = -uu_x + 0.1 u_{xx}$ |
| Clean data | $C_t = -1.000 C_x + 0.250 C_{xx}$ | $u_t = -0.997 uu_x + 0.100 u_{xx}$ |
| 1% noise | $C_t = -1.000 C_x + 0.250 C_{xx}$ | $u_t = -0.998 uu_x + 0.100 u_{xx}$ |
| 5% noise | $C_t = -0.996 C_x + 0.248 C_{xx}$ | $u_t = -0.963 uu_x + 0.096 u_{xx}$ |

From the table, it can be seen that, in engineering settings, DL-PDE can learn the PDE form and the corresponding coefficients accurately. The situation of fixed spatial observation points is also tested similarly, and satisfactory results are also obtained, which are provided in Appendix B. Previous methods are unsuitable for such problems because they rely on the finite difference method to find the derivatives, but temporal or spatial points here are discrete. It is worth mentioning that we find that, in engineering settings, DL-PDE is robust to measurement noise. For the case of slight noise (e.g., 1% noise), DL-PDE performs almost the same as in the case of no noise. Moreover, DL-PDE behaves well in the case of 5% noise.

### C. Comparison with other methods

#### 1. Comparison with direct STRidge

In DL-PDE, a deep neural network is used to represent physical processes, and meta-data are generated by this neural network to perform sparse regression. Compared with direct STRidge, which calculates the derivatives numerically based on the actual observation data, DL-PDE relies on deep-learning to come up with the derivatives via automatic differentiation. The direct STRidge requires the dataset to be on a regular grid, which means that data must be distributed uniformly in space and time. In contrast, DL-PDE is more flexible because it can handle discrete data points. In addition, when data are limited and noisy, the accuracy of derivatives in the direct STRidge, which is calculated by finite difference, is seriously affected while DL-PDE is robust to noise. Additional details are provided in Appendix C, in which the performance of DL-PDE and direct STRidge with data on a regular grid is compared.

#### 2. Comparison with DL-PDE without generating meta-data

In DL-PDE, meta-data are generated with the trained neural network. It is significant for PDE learning accuracy with discrete and noisy data. When the amount of observation data is small, the generated meta-data can assist to extract the essential features of the physical problem response, and can then benefit PDE learning. It is able to improve the stability of DL-PDE by generating a large number of meta-data to perform sparse regression. To investigate this, we compare the performance of DL-PDE with and without generating meta-data for different cases. In DL-PDE without generating meta-data, we employ the trained neural network to calculate the derivatives of each point on the original dataset without generating meta-data, and then utilize the STRidge to learn the PDE. We select the groundwater flow equation and the Burgers equation for illustration, and utilize these two methods to learn their corresponding PDEs from limited and noisy data. Conditions are the same as in previous cases. The results are presented in Table IX.

It can be seen from Table IX that the stability of DL-PDE can be augmented by generating a large amount of meta-data. When the noise level increases, DL-PDE with generating meta-data performs better, and the coefficients are more accurate. In the groundwater flow equation, whose original dataset is much smaller than meta-data, when meta-data are not generated, even if the form of the equation can be learned, the accuracy of the coefficient is very low in the presence of noise. Therefore, it is shown that generating meta-data is critical to improve the stability of DL-PDE.

**TABLE IX.** Comparison of the performance of DL-PDE with or without generating meta-data.

| | DL-PDE with Meta-data | DL-PDE without Meta-data |
|---|---|---|
| **1. Burgers equation (3000 data used)** | | |
| **Correct PDE** | $u_t = -uu_x + 0.1u_{xx}$ | |
| **Clean data** | $u_t = -0.996uu_x + 0.099u_{xx}$ | $u_t = -0.993uu_x + 0.099u_{xx}$ |
| **1% noise** | $u_t = -0.993uu_x + 0.098u_{xx}$ | $u_t = -0.993uu_x + 0.098u_{xx}$ |
| **5% noise** | $u_t = -0.986uu_x + 0.095u_{xx}$ | $u_t = -0.975uu_x + 0.094u_{xx}$ |
| **2. Groundwater flow equation (2500 data used)** | | |
| **Correct PDE** | $h^*_{t^*} = h^*_{x^*x^*}$ | |
| **Clean data** | $h^*_{t^*} = 0.992 h^*_{x^*x^*}$ | $h^*_{t^*} = 1.110 h^*_{x^*x^*}$ |
| **1% noise** | $h^*_{t^*} = 0.971 h^*_{x^*x^*}$ | $h^*_{t^*} = 0.026 h^*_{x^*x^*}$ |
| **5% noise** | $h^*_{t^*} = 0.906 h^*_{x^*x^*}$ | $h^*_{t^*} = 0.00024 h^*_{x^*x^*}$ |

## IV. SUMMARY AND DISCUSSION

Data-driven discovery provides a viable option for finding the unknown PDE of physical problems. However, many difficulties remain to be resolved prior to achieving this. In reality, the observation data are usually noisy and limited, which constitutes a major challenge for data-driven PDE discovery. To overcome this challenge, in this study, we proposed a novel method, called DL-PDE, combining deep neural network and sparse regression methods, such as Lasso, STRidge, and sparse Bayesian inference to identify hidden physical process and discover the corresponding governing equations. Compared to extant sparse regression methods for PDE discovery, it is not necessary for DL-PDE to use numerical differentiation to calculate derivatives, but instead automatic differentiation is utilized to generate derivatives via the neural network trained with the observation data. In addition, different from previous works that utilized neural networks, in this work, meta-data and the corresponding derivatives are generated by the trained neural network. The use of the meta-data is essential in improving the stability of the sparse regression algorithm, which results in learning the PDE form and the corresponding coefficients more accurately. In this approach, after representing the data accurately with the trained neural network, the underlying physical process can be expressed with a parsimonious model in terms of partial differential equations. Compared to the neural networks alone, interpretability, generalizability, and expandability are substantially augmented. In addition, compared to the direct sparse regression methods, problems with limited and noisy data are alleviated, and thus performance is significantly improved.

These assertions are confirmed with demonstrative examples and sensitivity studies. The numerical experiments show that DL-PDE is robust to data noise and performs well with sparse data. Moreover, without smoothing the noisy data, it can also learn the equation form and coefficients accurately. This is mainly because the automatic differentiation of the neural network is less affected by noise and possesses certain robustness to noise. The use of a large number of meta-data generated by trained neural networks also improves the accuracy of the STRidge process.

We have also discussed the performance of DL-PDE in actual engineering settings. Experiments have demonstrated that, in the case of fixed temporal or spatial monitoring points, DL-PDE works very well. This indicates that DL-PDE may be especially suitable for practical applications in which data are usually limited and corrupted with noise.

Finally, compared with direct STRidge, DL-PDE offers the advantages of dealing with sparse discrete data and is more robust to noise. In addition, numerical examples confirm that generating meta-data can greatly enhance the stability and accuracy of DL-PDE.

DL-PDE, at present, is not without limitations. The coefficients in the underlying governing equations are assumed to be constant in space and time. Recent developments in the literature that deal with smoothly varying coefficients [15] or piecewise-constant coefficients [7] may be incorporated in DL-PDE. However, the general case of spatially (or temporally) randomly distributed coefficients remains a challenge. The recent approach of combining sparse regression with data assimilation [7] that was devised for handling coefficients appearing as a nonlinear function of dependent variables with unknown parameters may prove to be useful for this endeavor, especially after incorporating the deep learning component proposed in this work.

In this work, well-known PDEs are utilized for proof-of-concept. In the future, experimental data of complex problems without knowing their inherent PDEs will be investigated utilizing the proposed method.

## ACKNOWLEDGEMENTS

This work is partially funded by the National Natural Science Foundation of China (Grant No. 51520105005 and U1663208), and the National Science and Technology Major Project of China (Grant No. 2017ZX05009-005 and 2017ZX05049-003).

# APPENDIX A: ILLUSTRATION OF SELECTED DATASET

In this appendix, we illustrate the selected dataset with different volume. The selected dataset for the groundwater flow equation, convection-diffusion equation, Burgers equation, and KdV equation are illustrated in Fig. A1, Fig. A2, Fig. A3, and Fig. A4, respectively.

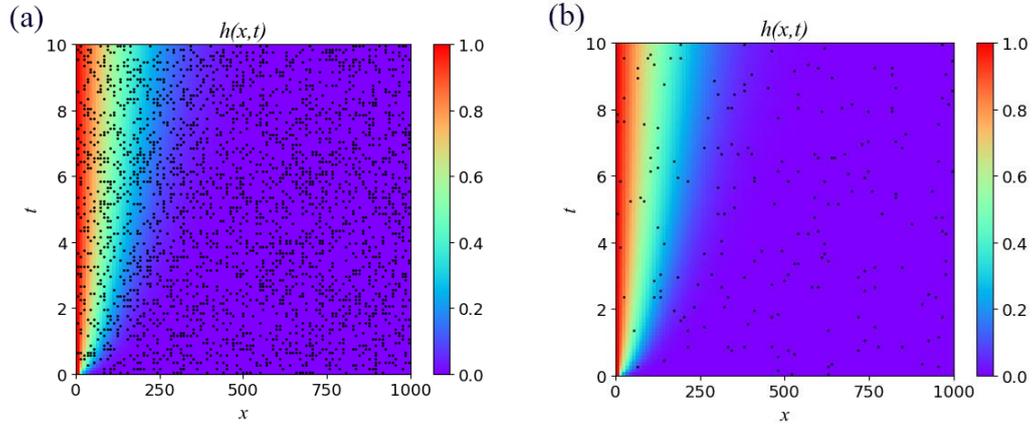

**FIG. A1.** Randomly selected data from the dataset of the groundwater flow equation: 2500 data (a) and 100 data (b). The background represents the solution $h$ in the dataset by heat map, and the black dots are selected data.

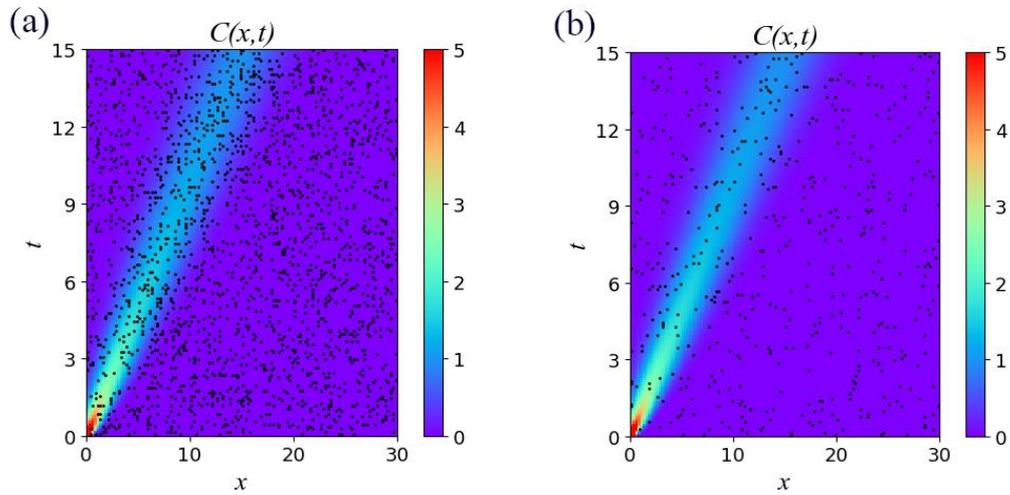

**FIG. A2.** Randomly selected data from the dataset of the convection-diffusion equation: 2000 data (a) and 500 data (b). The background represents the solution $C$ in the dataset by heat map, and the black dots are selected data.

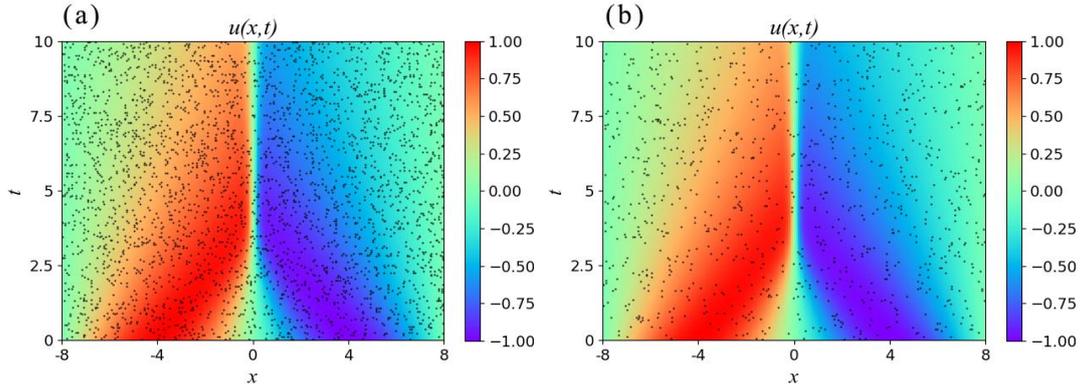

**FIG. A3.** Randomly selected data from the dataset of the Burgers equation: 3000 data (a) and 1000 data (b). The background represents the solution $u$ in the dataset by heat map, and the black dots are selected data.

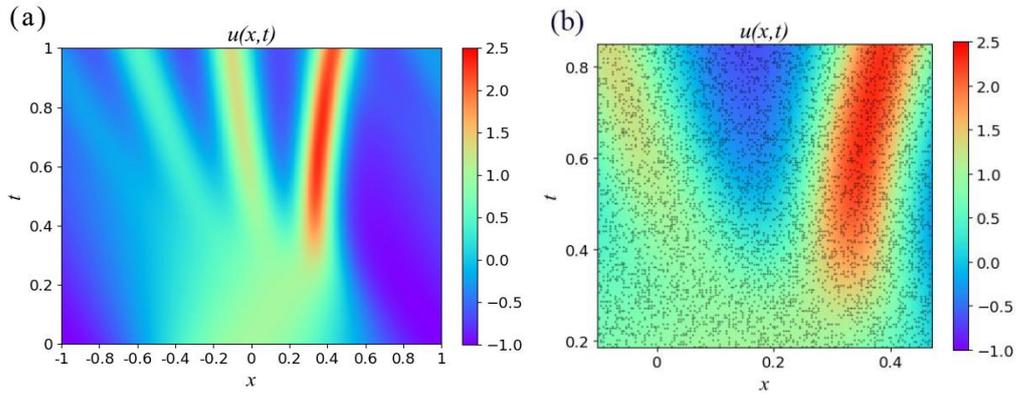

**FIG. A4.** The solution of the KdV equation in the form of heat map (a), and a portion of the map indicating 25000 data randomly selected from the dataset (b). The background represents the solution $u$ in the dataset by heat map, and the black dots are selected data.

## APPENDIX B: DISCOVERY OF PDES IN ENGINEERING SETTINGS WITH FIXED SPATIAL OBSERVATION POINTS

We use the convection-diffusion equation and the Burgers equation to conduct our experiments with fixed spatial observation locations. For the contaminant transport equation, a total of 12 spatial observation points are uniformly distributed in the entire domain. For the Burgers equation, a total of 50 spatial observation points are uniformly placed from $x=-7.9375$ to $x=7.5625$. We select the data at all temporal points at these spatial points as the dataset. Selected data are presented in Fig. B1.

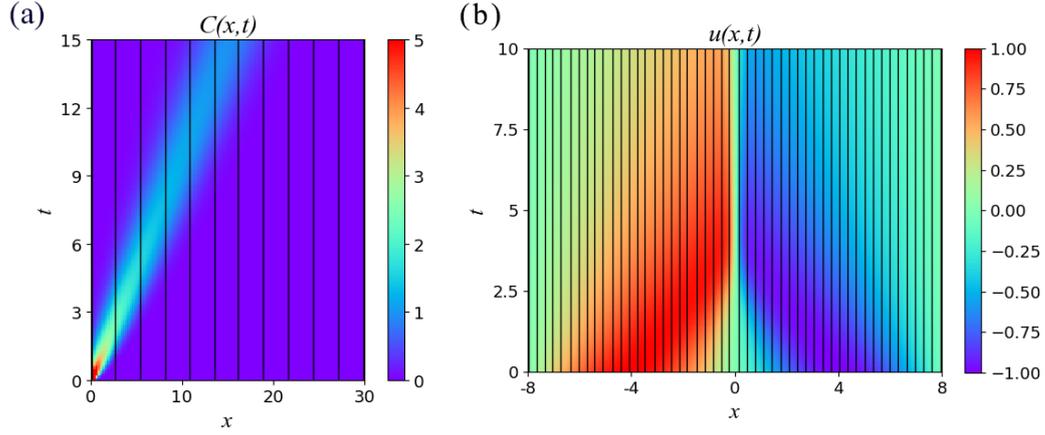

**FIG. B1.** Generating data from fixed spatial observation points. Training data are generated from 12 spatial observation locations in the case of convection-diffusion equation (a), and from 50 spatial observation locations in the case of Burgers equation (b). The background represents the solution $u$ in the dataset by heat map, and the black lines are selected data.

A five-layer deep neural network with 50 neurons per hidden layer is utilized to represent the solution $u$, and the activation functions used are sin(x). Meta-data are generated in the same way as in the previous corresponding examples. The results are presented in Table B I.

**TABLE B I.** Summary of PDEs learned using DL-PDE for the case of fixed spatial observation data.

| Noise Level | Learned Equation (convection diffusion equation) | Learned Equation (Burgers equation) |
|---|---|---|
| **Correct PDE** | $C_t = -C_x + 0.25 C_{xx}$ | $u_t = -u u_x + 0.1 u_{xx}$ |
| **Clean data** | $C_t = -1.000 C_x + 0.246 C_{xx}$ | $u_t = -0.990 u u_x + 0.100 u_{xx}$ |
| **1% noise** | $C_t = -1.000 C_x + 0.246 C_{xx}$ | $u_t = -0.990 u u_x + 0.101 u_{xx}$ |
| **5% noise** | $C_t = -1.004 C_x + 0.246 C_{xx}$ | $u_t = -0.973 u u_x + 0.097 u_{xx}$ |

**APPENDIX C: COMPARISON OF DL-PDE AND DIRECT STRIDGE**

Firstly, we use the DL-PDE and direct STRidge methods to discover the Burgers equation. In order to compare the performance of these two methods, we use the data points on the same regular grid as the dataset. For this case, we take spatial data points with $\Delta x = 0.0625$ in the domain $x \in [-8,8)$, and take temporal data points with $\Delta t = 0.05$ from $t=0$ to $t=10$. Consequently, we have $n_x$=256, $n_t$=201, and $N_d$=51456. In order to ensure that the data points are on a regular grid, we reduce the amount of data by selecting one observation point every few observation points in time or space. We use these two methods to learn the Burgers equation with 51456, 12928, and 3264 data, respectively. To obtain 12928 data, we take spatial points with $\Delta x = 0.125$ in the domain $x \in [-8,8)$ and take temporal points with $\Delta t = 0.1$ from $t=0$ to $t=10$, and we have a subset with $n_x$=128, $n_t$=101, and $N_d$=12928. To obtain 3264 data, we take spatial points with $\Delta x = 0.25$ in the domain $x \in [-8,8)$ and take temporal points with $\Delta t = 0.2$ from $t=0$ to $t=10$, and we have a subset with $n_x$=64, $n_t$=51, and $N_d$=3264. Selected data are displayed in Fig. C1.

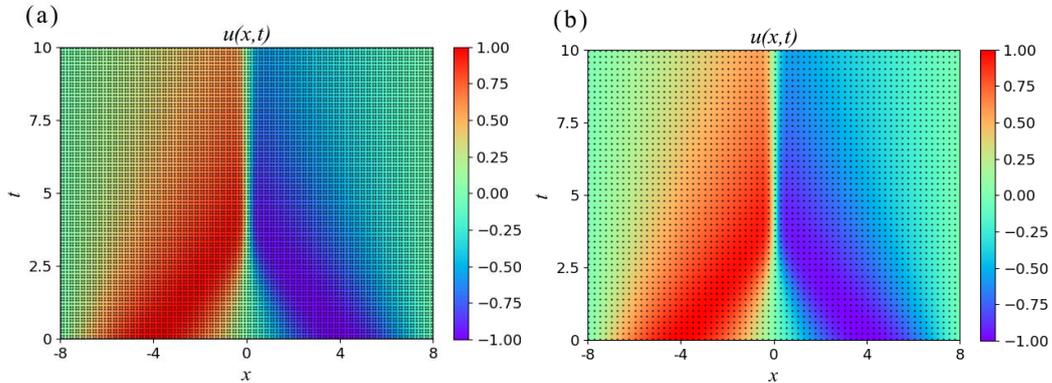

**FIG. C1.** Selected data from the dataset: 12928 data (a) and 3264 data (b). The background represents the solution $u$ in the dataset by heat map, and the black dots are selected data.

DL-PDE and direct STRidge are then utilized to learn the PDE, respectively. Outputs are shown in Table C I. It can be seen that, for the Burgers equation, the performances of these two methods are similar in the case of a large data volume, but when the data volume is small, the accuracy of direct STRidge is significantly reduced, whereas DL-PDE is still relatively more accurate. This shows that DL-PDE possesses obvious advantages over the direct STRidge for small data volume.

TABLE C I. Comparison of the performance of the two methods to learn the Burgers equation using different data volume.

|  | DL-PDE | Direct STRidge |
|---|---|---|
| Correct PDE | $u_t = -uu_x + 0.1u_{xx}$ | |
| 51456 data | $u_t = -0.998uu_x + 0.099u_{xx}$ | $u_t = -1.001uu_x + 0.102u_{xx}$ |
| 12928 data | $u_t = -0.997uu_x + 0.099u_{xx}$ | $u_t = -1.004uu_x + 0.109u_{xx}$ |
| 3264 data | $u_t = -0.989uu_x + 0.095u_{xx}$ | $u_t = -1.007uu_x + 0.134u_{xx}$ |

We next compare the performance of these two methods in the presence of noise. In DL-PDE, a nine-layer deep neural network with 20 neurons per hidden layer is trained to calculate derivatives. All data are used to train the neural network. In direct STRidge, a polynomial technique is utilized for smoothing noisy data to obtain stable and relatively accurate derivatives.

The results are displayed in Table C II. It can be seen that, in the absence of noise, the performance of the two methods is similar, but as the noise level increases, even if the noise has already been smoothed, the accuracy of direct STRidge decreases rapidly. Indeed, at the 5% noise level, it is unable to learn the correct equation form. In contrast, the performance of DL-PDE is very stable, and is robust to the 5% noise level with high accuracy. This indicates that DL-PDE is superior to direct STRidge in the presence of noise.

**TABLE C II.** Comparison of the performance of the two methods to learn the Burgers equation in the presence of noise.

|  | DL-PDE | Direct STRidge |
|---|---|---|
| **Correct PDE** | $u_t = -uu_x + 0.1u_{xx}$ | |
| **Clean data** | $u_t = -0.998uu_x + 0.099u_{xx}$ | $u_t = -1.001uu_x + 0.102u_{xx}$ |
| **1% noise** | $u_t = -0.994uu_x + 0.099u_{xx}$ | $u_t = -0.960uu_x + 0.102u_{xx}$ |
| **5% noise** | $u_t = -0.992uu_x + 0.098u_{xx}$ | $u_t = -0.824uu_x + 0.117u_{xx} - 0.06u^2u_{xx}$ |